\documentclass[10pt,letterpaper]{article}

\usepackage{cogsci}
\usepackage{booktabs}
\usepackage{graphicx}
\usepackage{multirow}
\usepackage{booktabs}
\usepackage{tabularx}
\usepackage{booktabs}
\usepackage{amsmath}

\cogscifinalcopy % Uncomment this line for the final submission 

\usepackage[
    backend=biber,
    style=apa,
    natbib=true,
    doi=false,
    isbn=false,
    url=false,
]{biblatex}
\usepackage{pslatex}
\usepackage{float} % Roger Levy added this and changed figure/table
\title{ChallengeMe: An Adversarial Learning-enabled Text Summarization Framework} 
\author{{\large \bf Xiaoyu Deng (xdeng24@fordham.edu)} New York, USA
\AND {\large \bf Ye Zhang (yez12@pitt.edu)}
Pennsylvania, USA
\AND {\large \bf Tianmin Guo (tg2374@nyu.edu)} 
  New York, USA
\AND {\large \bf Yongzhe Zhang (yongzhe@caltech.edu)} 
  California, USA
\AND {\large \bf Zhengjian Kang (zk299@nyu.edu)} 
  New York, USA
\AND {\large \bf Xintao Li (xintao.li@miami.edu)} 
  Florida, USA
  \AND {\large \bf Hang Yang (hyang113@terpmail.umd.edu)} 
  Maryland, USA
}
\begin{document}
\maketitle
\begin{abstract}
The astonishing performance of large language models (LLMs) and their remarkable achievements in production and daily life have led to their widespread application in collaborative tasks. However, current large models face challenges such as hallucination and lack of specificity in content generation in vertical domain tasks. Inspired by the contrast and classification mechanisms in human cognitive processes, this paper constructs an adversarial learning-based prompt framework named ChallengeMe, which includes three cascaded solutions: generation prompts, evaluation prompts, and feedback optimization. In this process, we designed seven core optimization dimensions and set the threshold for adversarial learning. The results of mixed case studies on the text summarization task show that the proposed framework can generate more accurate and fluent text summaries compared to the current advanced mainstream LLMs. 

\end{abstract}

\section{Introduction}

The development of artificial intelligence stems from researchers' contemplation and exploration of human-biological machines \citep{cheok2016love}. Inspired by the characteristics of biological appearance \citep{panagoulias2024evaluating}, living habits \citep{barthlott2016superhydrophobic}, and cognitive habits\citep{hertel2010cognitive}, a large number of bionic research designs and innovations have provided valuable explorations for the development of artificial intelligence. Among them, natural language processing technology aims to simulate human communication habits and construct human-like agents, which have extensive applications in fields such as question-answering systems. In recent years, with explosive growth of computing power and data, the development of large language models has been significant, showing superior performance and strong generalization and scalability in fields such as intelligent agents \citep{wooldridge1999intelligent}, service systems\citep{beuren2013product}, and healthcare \citep{daniels2001justice}, thereby significantly improving the efficiency of production and life.

Although general-purpose large language models (LLMs) have demonstrated astonishing performance, these models often perform poorly in vertical domain tasks, facing challenges such as generating nonspecific content \citep{pillitteri2012antidotal}, ignoring key information in prompts \citep{khurana2024and}, and severe hallucination issues \citep{rawte2023troubling}. In response to these challenges, many studies have explored optimization solutions, among which two representative strategies are the introduction of retrieval-augmented technology and fine-tuning technology. For instance, Li utilizes RAG to add an external knowledge domain to the model to reduce hallucinations \citep{li2024enhancing}, achieving an effect of efficiency. Guo adopts an adaptive fine-tuning strategy to enhance the model performance in specific domains \citep{guo2019spottune}, achieving an effect of high accuracy. However, these solutions have limitations: for strategies empowered by enhanced retrieval, the model performance is easily constrained by the scale of the external knowledge base and the training data \citep{jin2023improving}; for fine-tuning solutions, small-scale and significantly biased proprietary datasets can easily lead to a significant decline in the generalization performance of large models and hallucination issues \citep{zhang2024generalization}.

To address the aforementioned challenges, this paper, inspired by two typical behaviors in the human learning process: classification and comparison \citep{papini2002pattern}, designs a contrastive learning prompt strategy for large language models, thus constructing an adversarial prompt learning framework for text summarization tasks: \textit{ChallengeMe}. By embedding two prompt branches, content generation and content verification, into large language models and combining them with a feedback-enhanced threshold analysis strategy, high-quality text summarization performance is achieved. Compared with existing advanced solutions, the proposed solution in this paper has achieved superior performance in both objective indicators such as quality, fluency, and stability of generated summary content and subjective evaluation indicators from 30 human participants. Overall, the contributions of this paper are as follows:

This paper constructs an adversarial learning prompt framework based on the contrastive learning approach in human cognitive processes. By adopting a dual-branch scheme of a generator and an inspector, combined with a threshold feedback calibration scheme, high-quality text summarization performance is achieved. 

This paper conducts mixed experiments to verify the effectiveness of the proposed framework from both quantitative and qualitative perspectives. 

This paper discusses and analyzes the conclusions and findings of the experiments, proposing potential directions and ideas for the optimization of future human-like large models. 

The structure of the paper is as follows: Chapter 2 outlines the current research status in the field, Chapter 3 introduces the research methods used in this paper, Chapter 4 describes the experimental implementation details and results, Chapter 5 analyzes the experiments and conducts discussions.

\section{Related Works}

The cognitive characteristics of achieving adversarial learning through classification and comparison are not rare in nature, such as in primates \citep{firestone2020performance}, cetaceans \citep{connor2006social}, and humans\citep{firestone2020performance}. With the significant progress made by Wang in introducing adversarial learning into the field of computer science \citep{wang2017generative}, there has now been a wealth of exploratory practices in research on adversarial sample learning. In the area of computer vision, adversarial learning has primarily been used to enhance model robustness against adversarial attacks. For instance, Goyal introduced adversarial training as a defense mechanism, where models are trained on adversarial examples to improve their robustness \citep{goyal2023survey}. Meanwhile, Goyal’s framework has been extended to improve generative tasks, such as image synthesis and style transfer, leveraging frameworks like Generative Adversarial Networks (GANs) \citep{goodfellow2020generative}. For example, Su demonstrated the utility of adversarial examples in domain adaptation tasks \citep{su2020active}, and the result indicates that such method enables models to generalize better across diverse datasets. Adversarial learning has also been widely adopted in natural language processing (NLP) tasks. For example, Zhang explored adversarial attacks to reveal vulnerabilities in text classification models and achieved great performance \citep{zhang2020adversarial}. Beyond attacks, adversarial learning has been employed to improve the generalization and robustness of NLP models. For instance, adversarial data augmentation has been utilized to train models that are less sensitive to noise in input text. Moreover, adversarial robust methods have been developed to mitigate biases in text generation and sentiment analysis tasks, as Textbugger proposed by Li et al. had demonstrated great performance on text generation tasks.

With the widespread application of Large Language Models (LLMs), some studies have indicated that LLMs possess the potential to pass the Turing Test \citep{sejnowski2023large}. Even more optimistically, certain research suggests that LLMs have, to a certain extent, already passed the Turing Test \citep{sejnowski2023large}. However, the current cognitive level of LLMs still lags significantly behind that of humans and has considerable room for improvement. Therefore, many studies explore the differences between AI and humans, hoping to conduct comparative research through these differences to promote the development of AI \citep{djeffal2022role}. Existing studies have sought to compare AI and human information processing from a cognitive science perspective and obtain valuable insights. For instance, recent work has focused on the discrepancies between LLMs and humans in reasoning tasks, particularly in commonsense reasoning and contextual understanding. These studies often design adversarial datasets or specific cognitive tasks to evaluate the logical consistency and generalization capabilities of LLMs \citep{ying2023intuitive, gu2024survey}. Such comparative research not only highlights the weaknesses of AI models but also offers guidance for their optimization \citep{surianarayanan2023survey}. Second, some research efforts have drawn inspiration from human learning mechanisms to enhance AI models. Techniques like meta-learning and analogy-based reasoning have been applied to improve the performance of LLMs in few-shot learning tasks \citep{yuan2023analogykb}. Humans' ability to reason by analogy has inspired researchers to integrate this cognitive trait into LLMs, significantly boosting their performance in low-resource or unfamiliar domains \citep{guo2024teaching}. Additionally, other work has focused on addressing the differences between AI and humans in handling ambiguity and uncertainty, leading to the development of more robust model architectures \citep{chander2024toward}. Overall, these works demonstrate that leveraging insights from the differences between AI and humans can not only drive the advancement of AI models but also enhance their value in real-world applications. This research builds upon this foundation by introducing a bidirectional improvement mechanism inspired by generative adversarial networks (GANs), simulating ``mutual learning" between AI and humans to optimize the generation capabilities of the model.

\section{Method}

\subsection{Overall Design}

To achieve high-quality text summarization tasks, we mimic the classification and comparison features in the human learning process to construct ChallengeMe, an adversarial prompt-driven text summarization framework. Unlike existing prompt schemes, this framework adopts a multi-round optimization strategy to enhance the consistency and fluency of the output content. The proposed framework consists of three modules, namely input prompts, adversarial prompts, and feedback optimization strategies, as shown in the figure below.
\begin{figure*}
\centering
    \includegraphics[width=0.85\linewidth]{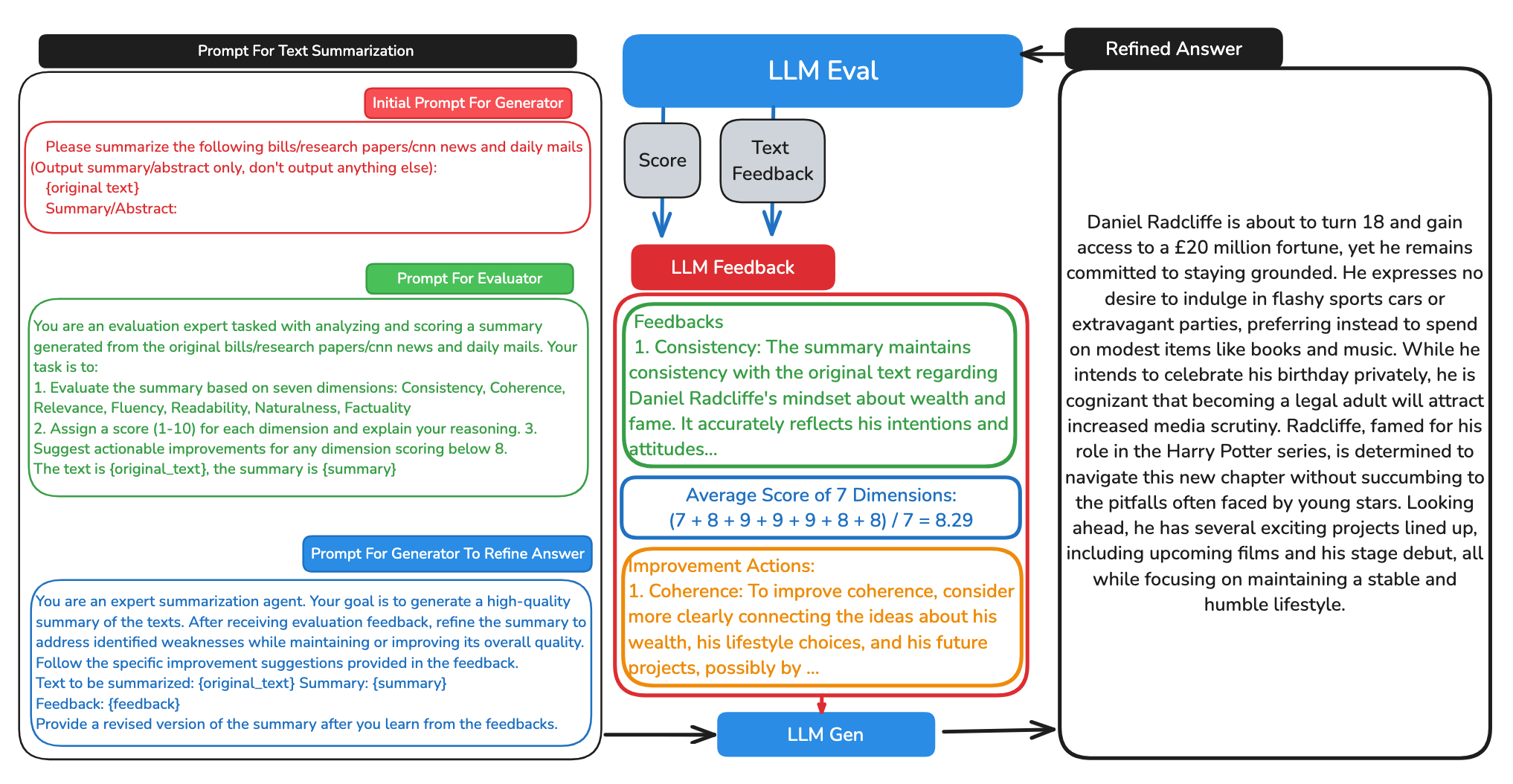}
    \caption{Fig. Framework of the proposed ChallengeMe}
\end{figure*}

\subsection{Framework }
\subsubsection{\textbf{Prompt Generation} }

This paper proposes a generation framework based on adversarial prompts. The core idea of the framework is to guide the LLM to follow specific goals and constraints during text summarization by designing appropriate prompts. First, the model needs to accept the input text:

\begin{subequations}
\begin{align}
T_{in} = { t_1, t_2, \dots, t_n }
\end{align}
\end{subequations}

where $t_i$ represents the i-th word in the text, and the text length is \textit{n}. The prompt part guides the model through explicit task requirements. On this basis, we define the summarization target as:

\begin{subequations}
\begin{align}
{Prompt}_{target} = \\ \notag
{``\textit{Summarize the following text to highlight key points:"}}
\end{align}
\end{subequations}

% ${Prompt}_{target} =\\ {``\textit{Summarize the following text to highlight key points:"}}$

This prompt statement clearly instructs the model to extract key information from the text for summarization. In addition, to ensure that the generated summary is concise, we add a summary length constraint to the prompt, setting the maximum number of words in the summary to \textit{L}max:

% $Prompt_{length}=\\{``\textit{The summary should not exceed "}}$

\begin{subequations}
\begin{align}
Prompt_{length}= \\ \notag
{``\textit{The summary should not exceed "}}
\end{align}
\end{subequations}

Furthermore, this study sets a language style constraint for the model-generated text, and in this paper, the target style is a concise and formal language style:

\begin{subequations}
\begin{align}
{Prompt}_{length}= \\ \notag
 {``\textit{The summary should not exceed }} L_{{max}} { words."}
\end{align}
\end{subequations}

% ${Prompt}_{length} =\\ {``\textit{The summary should not exceed }} L_{{max}} { words."}$

The final generated prompt is as follows:

\begin{subequations}
\begin{align}
{Prompt} = {Prompt}_{target} + {Prompt}_{length} + {Prompt}_{style}
\end{align}
\end{subequations}

% ${Prompt} = {Prompt}_{target} + {Prompt}_{length} + {Prompt}_{style}$

Under this prompt framework, the model generates the summary \textit{S}out based on the input text \textit{T}in and the prompt. The specific generation process can be represented as:

% $S_{out} = g(f(T_{in}), Prompt)$

\begin{subequations}
\begin{align}
S_{out} = g(f(T_{in}), Prompt)
\end{align}
\end{subequations}

where $f(T_{in}$ represents the encoding of the input text \textit{T}in by the model, and the obtained representation is used to guide the decoder to generate the summary, while $\textit{g}()$ is the generation process of the decoder.

\subsubsection{Detect the Prompt}
To ensure the quality of the generated summaries, this paper designs a prompt detector to evaluate from multiple dimensions: Consistency, Coherence, Relevance, Fluency, Readability, Naturalness, and Factuality, with the score range for each evaluation dimension being 1 to 10.

For fluency, this paper defines it as $\textit{F}_{fluency}$, which is used to assess whether the language of the generated text is natural and smooth. The fluency score is calculated by the following formula:

\begin{subequations}
\begin{align}
F_{fluency} = \frac{1}{m} \sum_{i=1}^{m} {fluency}(S_{out}^i)
\end{align}
\end{subequations}

% $F_{fluency} = \frac{1}{m} \sum_{i=1}^{m} {fluency}(S_{out}^i)$

where \textit{S}out\textit{i} represents the \textit{i}-th sentence in the summary, fluency(\textit{S}out\textit{i}) is the fluency score of that sentence, and \textit{m} is the number of sentences in the summary.

For Consistency, this paper defines it as $\textit{F}_{consistency}$, which is used to assess whether the summary accurately reflects the key information of the original text. The consistency score is calculated by comparing the overlap between the generated summary and the original text, with the formula being:

\begin{subequations}
\begin{align}
F_{consistency} = \frac{| S_{out} \cap T_{in} |}{| T_{in} |}
\end{align}
\end{subequations}

% $F_{accuracy} = \frac{| S_{out} \cap T_{in} |}{| T_{in} |}$

where $| S_{out} \cap T_{in} |$ is the number of overlapping words between the generated summary and the input text, and $| T_{in} |$is the total number of words in the original text.

For language Naturalness, this paper defines it as $\textit{F}_{naturalness}$, which is used to assess whether the language expression of the generated summary is concise and elegant. Its score calculation formula is:

\begin{subequations}
\begin{align}
F_{naturalness} = \frac{1}{m} \sum_{i=1}^{m} {naturalness}(S_{out}^i)
\end{align}
\end{subequations}

% $F_{elegance} = \frac{1}{m} \sum_{i=1}^{m} {elegance}(S_{out}^i)$

where naturalness(\textit{S}out\textit{i}) is the naturalness score of the \textit{i}-th sentence, and \textit{m} is the number of sentences in the summary.

To determine whether the generated summary meets the requirements, the detection model will give scores according to the above three dimensions. Let the scores for each dimension be $\textit{F}_{fluency}$, $\textit{F}_{consistency}$, and $\textit{F}_{naturalness}$, the detection model will give a score for each dimension, with the score range being 1 to 10. Only when the scores for all evaluation dimensions meet the minimum threshold will the output of the generation model be accepted as a valid summary. Specifically, assuming that the threshold for each dimension is \textit{T}min=7, the output \textit{S}out of the generation model will only be accepted under the following conditions:

% $F_{fluency} \geq T_{min}, \quad F_{accuracy} \geq T_{min}, \quad F_{elegance} \geq T_{min}$

\begin{subequations}
\begin{align}
F_{fluency} \geq T_{min}, \quad F_{consistency} \geq T_{min}, \quad F_{naturalness} \geq T_{min}
\end{align}
\end{subequations}

\subsection{Feedback Optimization Strategy}
To further enhance the quality of the generated summaries, we have designed an adaptive feedback optimization strategy. When the detection model evaluates the output of the generation model, if the score in a certain dimension does not reach the preset minimum threshold, the generation model will adjust itself based on the feedback to optimize the generation process. This section takes the parameter control of three dimensions: fluency, consistency, and naturalness as examples to conduct analysis:

If the fluency score $F_{fluency}$ is lower than 7, the generation model will receive feedback prompts to improve the sentence structure of the summary, making it more natural and fluent:

% ${Prompt}_{fluency} = \\{``\textit{Rewrite the summary with more natural sentence structures."}}$

\begin{subequations}
\begin{align}
{Prompt}_{fluency} = \\ \notag
{``\textit{Rewrite the summary with more natural sentence structures."}}
\end{align}
\end{subequations}

If the consistency score $\textit{F}_{consistency}$ is lower than 7, the generation model will be prompted to increase the coverage of key information to ensure the summary is more accurate:

% ${Prompt}_{accuracy} = \\{``\textit{Ensure all key points are included in the summary."}}$

\begin{subequations}
\begin{align}
{Prompt}_{consistency} = \\ \notag
{``\textit{Ensure all key points are included in the summary."}}
\end{align}
\end{subequations}

If the language naturalness score $\textit{F}_{naturalness}$ is lower than 7, the generation model will be prompted to optimize the language style to make the summary more concise and elegant:

% ${Prompt}_{elegance} = \\{``\textit{Make the language more elegant and concise."}}$

\begin{subequations}
\begin{align}
{Prompt}_{naturalness} = \\ \notag
{``\textit{Make the language more naturalness and concise."}}
\end{align}
\end{subequations}

Specifically, assuming $\Delta S_{out}$ is the adjustment of the summary generation process by the generation model after receiving feedback, the optimized output can be represented as:

% $S_{out}^{new} = S_{out} + \Delta S_{out}$

\begin{subequations}
\begin{align}
S_{out}^{new} = S_{out} + \Delta S_{out}
\end{align}
\end{subequations}

When the scores in all evaluation dimensions reach the minimum threshold, the final output content is generated.

\subsection{Formation of Prompts }

As shown in Figure 2, the prompt provided to the generation model (LLMgen) includes a scenario description, an SAP prompt, a list of actions, and an example plan. The SAP prompt aims to elicit complex reasoning by encouraging the model to thoroughly consider the different needs and potential interactions between people, animals, and objects. By explicitly prompting the model to infer the needs of other entities and predict how the situation might dynamically evolve, the prompt promotes empathy and holistic thinking, which are crucial for designing comprehensive plans. This one-time example illustrates the required plan structure in code format without providing solutions specific to the evaluation scenario. In contrast, the prompt for the evaluation model (LLMeval) includes an FSM plan generated by LLMgen, a benchmark high-quality plan, and a description of the scoring criteria for assessing plan quality through iterative improvement (see Appendix Figure 17). Initially, the benchmark plan includes manually written solutions, but in subsequent iterations, they include the highest-scoring automatically generated plans from the previous round. 

\section{Results }
\subsection{Settings }
During the experimental process, we utilized the Win/Mac operating systems and deployed/tested the prototype on Pycharm version 13.2. The study selected GPT-4o, Zhipu AI, Claude, Doubao, and Mistral-7b as baselines for comparative research with ChallengeMe. The aforementioned baselines are widely used general large models, and their performance stability and output result reliability have been extensively verified. Moreover, the performance levels of these models are all within the 4-4.5 generation versions (taking the GPT series as the standard), which can well represent the performance of the most advanced large models. Regarding the datasets, the experiment chose data sets from three sources: CNN/Daily Mail \citep{nallapati2016abstractive}, BillSum \citep{kornilova2019billsum}, and arXiv Summarization Dataset \citep{cohan2018discourse} to conduct quantitative experiments, ablation experiments, and qualitative experimental evaluations in order to enhance the reliability of the experimental results.
\subsection{Quantitative Evaluation  }
In the quantitative experiments, this paper selects Rouge1-Rouge5 \citep{ganesan2018rouge}, Rouge-L \citep{lin2004rouge}, BLEU \citep{papineni2002bleu}, Meteor \citep{banerjee2005meteor}, and Bertscore \citep{zhang2019bertscore} as evaluation metrics to assess model performance from multiple dimensions. The experimental results are shown in the table below. It can be seen that on the CNN/Daily Mail dataset, our model achieves the best performance in Rouge1, Rouge2, Rouge5, Bleu, and Bertscore metrics, and the second-best performance in Rouge4; on the BillSum dataset, the proposed model achieves the best performance in Bleu and Bertscore metrics, and the second-best performance in Rouge1, Rouge2, Rouge3, Rouge4, and Meteor metrics; on the arXiv Summarization Dataset, the proposed model achieves the best performance in Rouge1-4, Bleu, and Bertscore metrics, and the second-best performance in Rouge5, RougeL, and Meteor metrics. This indicates that the proposed solution in this paper has significant superiority in terms of content generation quality and stability compared to existing advanced baselines.

\begin{table}[h]
\setlength{\tabcolsep}{0.5pt}
    \centering
    \begin{tabular}{lccccccccc}
        \hline
        \textbf{} & \textbf{\scriptsize Rouge1} & \textbf{\scriptsize Rouge2} & \textbf{\scriptsize Rouge3} & \textbf{\scriptsize Rouge4} & \textbf{\scriptsize Rouge5} & \textbf{\scriptsize RougeL} & \textbf{\scriptsize Bleu} & \textbf{\scriptsize Metetor} & \textbf{\scriptsize Bertscore} \\ \hline
        \scriptsize GPT-4o & \scriptsize 0.2928 & \scriptsize 0.0897 & \scriptsize 0.0332 & \scriptsize 0.0117 & \scriptsize 0.0054 & \scriptsize 0.1795 & \scriptsize 0.0099 & \scriptsize 0.2491 & \scriptsize 0.8646 \\ 
        
        \scriptsize Zhipu AI &\scriptsize  0.2978 &\scriptsize  0.1163 & \textbf{\scriptsize 0.0563} & \scriptsize 0.0290 & \scriptsize 0.0170 & \textbf{\scriptsize 0.1968} & \scriptsize 0.0294 & \textbf{\scriptsize 0.3003} & \scriptsize 0.8648 \\ 
        
        \scriptsize Claude & \scriptsize 0.3157 & \scriptsize 0.0953 & \scriptsize 0.0334 & \scriptsize 0.0130 & \scriptsize 0.0051 & \scriptsize 0.1859 &\scriptsize  0.0104 & \scriptsize 0.2511 & \scriptsize 0.8645 \\ 
        
        \scriptsize Doubao & \scriptsize 0.3069 &\scriptsize  0.1006 & \scriptsize 0.0393 &\scriptsize  0.0195 & \scriptsize 0.0108 & \scriptsize 0.1932 & \scriptsize 0.0200 & \scriptsize 0.2577 & \scriptsize 0.8653 \\ 
        
        \scriptsize Mistral-7b & \scriptsize 0.2683 & \scriptsize 0.0952 & \scriptsize 0.0429 & \scriptsize 0.0220 & \scriptsize 0.0118 & \scriptsize 0.1712 &\scriptsize  0.0185 & \scriptsize 0.2753 & \scriptsize 0.8593 \\ 
        
        \scriptsize Ours & \textbf{\scriptsize 0.3188} & \textbf{\scriptsize 0.1290} & \scriptsize 0.0421 & \scriptsize 0.0268 & \textbf{\scriptsize 0.0197} & \scriptsize 0.1909 & \textbf{\scriptsize 0.0297} & \scriptsize 0.2915 & \textbf{\scriptsize 0.8704} \\ \hline
    \end{tabular}
    \caption{CNN/Daily Mail Quantitative Evaluation Metrics}
    \label{table:label_placeholder}
\end{table}

\begin{table}[h]
\setlength{\tabcolsep}{0.5pt}
    \centering
    \begin{tabular}{cccccccccc}
        \hline
        \textbf{} & \textbf{\scriptsize Rouge1} & \textbf{\scriptsize Rouge2} & \textbf{\scriptsize Rouge3} & \textbf{\scriptsize Rouge4} & \textbf{\scriptsize Rouge5} & \textbf{\scriptsize RougeL} & \textbf{\scriptsize Bleu} & \textbf{\scriptsize Metetor} & \textbf{\scriptsize Bertscore} \\ \hline
        \scriptsize GPT-4o & \scriptsize 0.4492 & \scriptsize 0.1874 &\scriptsize  0.1093 & \scriptsize 0.0722 & \scriptsize 0.0483 & \scriptsize 0.2834 & \scriptsize 0.0821 & \scriptsize 0.2850 & \scriptsize 0.8740 \\ 
        
        \scriptsize Zhipu AI & \scriptsize 0.4347 & \scriptsize 0.2243 & \scriptsize 0.1377 & \scriptsize 0.0857 & \scriptsize 0.0680 & \scriptsize 0.4043 & \scriptsize 0.1823 & \scriptsize 0.3045 & \scriptsize 0.8736 \\ 
        
        \scriptsize Claude & \scriptsize 0.4347 & \scriptsize 0.1790 & \scriptsize 0.1006 & \scriptsize 0.0616 & \scriptsize 0.0387 & \scriptsize 0.2793 & \scriptsize 0.0697 & \scriptsize 0.2667 & \scriptsize 0.8642 \\ 
        
        \scriptsize Doubao & \scriptsize 0.4295 & \scriptsize 0.2096 & \scriptsize 0.1365 & \scriptsize 0.0953 & \scriptsize 0.0670 & \scriptsize 0.3111 &\scriptsize  0.0910 & \scriptsize 0.2708 & \scriptsize 0.8682 \\ 
        
        \scriptsize Mistral-7b & \textbf{\scriptsize 0.4980} & \textbf{\scriptsize 0.2523} & \textbf{\scriptsize 0.1691} & \textbf{\scriptsize 0.1232} & \textbf{\scriptsize 0.0907} & \textbf{\scriptsize 0.3452} & \scriptsize 0.1388 & \textbf{\scriptsize 0.3549} & \scriptsize 0.8804 \\  
        \scriptsize Ours & \scriptsize 0.4948 & \scriptsize 0.2330 & \scriptsize 0.1597 & \scriptsize 0.1101 & \scriptsize 0.0819 & \scriptsize 0.3323 & \textbf{\scriptsize 0.1426} & \scriptsize 0.3232 & \textbf{\scriptsize 0.8894} \\ \hline
    \end{tabular}
    \caption{BillSum Quantitative Evaluation Metrics}
    \label{table:label_placeholder}
\end{table}

\begin{table}[h]
\setlength{\tabcolsep}{0.5pt}
    \centering
    \begin{tabular}{cccccccccc}
        \hline
        \textbf{} & \textbf{\scriptsize Rouge1} & \textbf{\scriptsize Rouge2} & \textbf{\scriptsize Rouge3} & \textbf{\scriptsize Rouge4} & \textbf{\scriptsize Rouge5} & \textbf{\scriptsize RougeL} & \textbf{\scriptsize Bleu} & \textbf{\scriptsize Metetor} & \textbf{\scriptsize Bertscore} \\ \hline
        \scriptsize GPT-4o &\scriptsize  0.4077 &\scriptsize  0.1284 &\scriptsize  0.0540 &\scriptsize  0.0263 & \scriptsize 0.0132 &\scriptsize  0.2179 &\scriptsize  0.0302 &\scriptsize  0.2177 & \scriptsize 0.8321 \\  
        
        \scriptsize Zhipu AI &\scriptsize  0.4159 & \scriptsize 0.1633 &\scriptsize  0.0758 & \scriptsize 0.0409 & \textbf{\scriptsize 0.0239} & \scriptsize 0.2356 & \scriptsize 0.0523 & \textbf{\scriptsize 0.2447} & \scriptsize 0.8339 \\  
        \scriptsize Claude &\scriptsize  0.4449 & \scriptsize 0.1644 & \scriptsize 0.0763 &\scriptsize  0.0387 & \scriptsize 0.0204 & \textbf{\scriptsize 0.2486} & \scriptsize 0.0467 & \scriptsize 0.2330 & \scriptsize 0.8384 \\  
       \scriptsize  Doubao &\scriptsize  0.2260 & \scriptsize 0.0609 &\scriptsize  0.0205 & \scriptsize 0.0069 & \scriptsize 0.0036 & \scriptsize 0.1734 & \scriptsize 0.0048 & \scriptsize 0.1056 & \textbf{\scriptsize 0.8481} \\  
        \scriptsize Mistral-7b & \scriptsize 0.3893 & \scriptsize 0.1485 &\scriptsize  0.0696 & \scriptsize 0.0372 & \scriptsize 0.0209 & \scriptsize 0.2165 & \scriptsize 0.0451 & \scriptsize 0.2397 & \scriptsize 0.8300 \\ 
        
        \scriptsize Ours & \textbf{\scriptsize 0.4575} & \textbf{\scriptsize 0.1659} & \textbf{\scriptsize 0.0797} & \textbf{\scriptsize 0.0454} & \scriptsize 0.0219 &\scriptsize  0.2363 & \textbf{\scriptsize 0.0526} & \scriptsize 0.2432 & \scriptsize 0.8394 \\ \hline
    \end{tabular}
    \caption{arXiv Summarization Dataset Quantitative Evaluation Metrics}
    \label{table:label_placeholder}
\end{table}
\subsection{Ablation Evaluation  }
To verify the rationality of the proposed framework and the effectiveness of the synergistic effect, we conducted ablation experiments for research. The discrimination threshold was set to 8.0--8.8 respectively, and experiments were carried out with a step length of 0.2 to verify the output performance of the model, and the results shown in the following table were obtained. It is not difficult to see from it that with the continuous increase of the threshold, the performance of the model gradually increases, and when the discrimination threshold exceeds 8.8, the performance of the output content tends to be stable, which indicates that the threshold (8.8) set by this solution can reduce the resource consumption on the basis of ensuring the quality of the output content.
\begin{table}[h]
\setlength{\tabcolsep}{0.5pt}
    \centering
    \begin{tabular}{cccccccccc}
        \hline
        \textbf{} & \textbf{\scriptsize Rouge1} & \textbf{\scriptsize Rouge2} & \textbf{\scriptsize Rouge3} & \textbf{\scriptsize Rouge4} & \textbf{\scriptsize Rouge5} & \textbf{\scriptsize RougeL} & \textbf{\scriptsize Bleu} & \textbf{\scriptsize Metetor} & \textbf{\scriptsize Bertscore} \\ \hline
\scriptsize TS=8.0 & \scriptsize 0.2647 & \scriptsize 0.0822 & \scriptsize 0.0301 & \scriptsize 0.0122 & \scriptsize 0.0057 & \scriptsize 0.1617 & \scriptsize 0.0164 & \scriptsize 0.2498 & \scriptsize 0.8600 \\  
\scriptsize TS=8.2 & \scriptsize 0.2710 & \scriptsize 0.0836 & \scriptsize 0.0293 & \scriptsize 0.0107 & \scriptsize 0.0031 & \scriptsize 0.1572 & \scriptsize 0.0159 & \scriptsize 0.2697 & \scriptsize 0.8610 \\  
\scriptsize TSd=8.4 & \scriptsize 0.2892 & \scriptsize 0.0934 & \scriptsize 0.0370 & \scriptsize 0.0140 & \scriptsize 0.0060 & \scriptsize 0.1660 & \scriptsize 0.0204 & \scriptsize 0.2859 & \scriptsize 0.8655 \\  
\scriptsize TS=8.6 & \scriptsize 0.2914 & \scriptsize 0.1037 & \scriptsize 0.0496 & \scriptsize 0.0267 & \scriptsize 0.0140 & \scriptsize 0.1758 & \scriptsize 0.0289 & \scriptsize 0.2943 & \scriptsize 0.8671 \\ 
\scriptsize TS=8.8 & \scriptsize 0.3188 & \scriptsize 0.1290 & \scriptsize 0.0421 & \scriptsize 0.0268 & \scriptsize 0.0197 & \scriptsize 0.1909 & \scriptsize 0.0297 & \scriptsize 0.2915 & \scriptsize 0.8704 \\ \hline
    \end{tabular}
    \caption{CNN/Daily Mail Quantitative Evaluation Metrics}
    \label{table:label_placeholder}
\end{table}

\begin{table}[h]
\setlength{\tabcolsep}{0.5pt}
    \centering
    \begin{tabular}{cccccccccc}
        \hline
        \textbf{} & \textbf{\scriptsize Rouge1} & \textbf{\scriptsize Rouge2} & \textbf{\scriptsize Rouge3} & \textbf{\scriptsize Rouge4} & \textbf{\scriptsize Rouge5} & \textbf{\scriptsize RougeL} & \textbf{\scriptsize Bleu} & \textbf{\scriptsize Metetor} & \textbf{\scriptsize Bertscore} \\ \hline
        \scriptsize TS=8.0 & \scriptsize 0.4388 & \scriptsize 0.1743 & \scriptsize 0.0951 & \scriptsize 0.0596 & \scriptsize 0.0374 & \scriptsize 0.2696 & \scriptsize 0.0717 & \scriptsize 0.2936 & \scriptsize 0.8694 \\ 
        \scriptsize TS=8.2 & \scriptsize 0.4465 & \scriptsize 0.1741 & \scriptsize 0.0950 & \scriptsize 0.0592 & \scriptsize 0.0371 & \scriptsize 0.2714 & \scriptsize 0.0728 & \scriptsize 0.2814 & \scriptsize 0.8698 \\ 
        \scriptsize TS=8.4 & \scriptsize 0.4559 & \scriptsize 0.1793 & \scriptsize 0.1027 & \scriptsize 0.0661 & \scriptsize 0.0431 & \scriptsize 0.2736 & \scriptsize 0.0815 & \scriptsize 0.3053 & \scriptsize 0.8717 \\ 
        \scriptsize TS=8.6 & \scriptsize 0.4738 & \scriptsize 0.1872 & \scriptsize 0.1063 & \scriptsize 0.0660 & \scriptsize 0.0418 & \scriptsize 0.2769 & \scriptsize 0.0861 & \scriptsize 0.3026 & \scriptsize 0.8683 \\ 
        \scriptsize TS=8.8 & \scriptsize 0.4948 & \scriptsize 0.2330 & \scriptsize 0.1597 & \scriptsize 0.1101 & \scriptsize 0.0819 & \scriptsize 0.3323 & \scriptsize 0.1426 & \scriptsize 0.3232 & \scriptsize 0.8894 \\ \hline
    \end{tabular}
    \caption{BillSum Quantitative Evaluation Metrics}
    \label{table:label_placeholder}
\end{table}
\begin{table}[h]
\setlength{\tabcolsep}{0.5pt}
    \centering
    \begin{tabular}{cccccccccc}
\hline
        \textbf{} & \textbf{\scriptsize Rouge1} & \textbf{\scriptsize Rouge2} & \textbf{\scriptsize Rouge3} & \textbf{\scriptsize Rouge4} & \textbf{\scriptsize Rouge5} & \textbf{\scriptsize RougeL} & \textbf{\scriptsize Bleu} & \textbf{\scriptsize Metetor} & \textbf{\scriptsize Bertscore} \\ \hline
        \scriptsize TS=8.0 & \scriptsize 0.4080 & \scriptsize 0.1246 & \scriptsize 0.0487 & \scriptsize 0.0206 & \scriptsize 0.0078 & \scriptsize 0.2176 & \scriptsize 0.0249 & \scriptsize 0.2278 & \scriptsize 0.8335 \\ 
        \scriptsize TS=8.2 & \scriptsize 0.4160 & \scriptsize 0.1321 & \scriptsize 0.0527 & \scriptsize 0.0245 & \scriptsize 0.0113 & \scriptsize 0.2138 & \scriptsize 0.0294 & \scriptsize 0.2254 & \scriptsize 0.8341 \\ 
        \scriptsize TS=8.4 & \scriptsize 0.4287 & \scriptsize 0.1365 & \scriptsize 0.0552 & \scriptsize 0.0259 & \scriptsize 0.0129 & \scriptsize 0.2219 & \scriptsize 0.0279 & \scriptsize 0.2286 & \scriptsize 0.8362 \\ 
        \scriptsize TS=8.6 & \scriptsize 0.4455 & \scriptsize 0.1423 & \scriptsize 0.0550 & \scriptsize 0.0242 & \scriptsize 0.0116 & \scriptsize 0.2269 & \scriptsize 0.0323 & \scriptsize 0.2337 & \scriptsize 0.8372 \\ 
        \scriptsize TS=8.8 & \scriptsize 0.4575 & \scriptsize 0.1659 & \scriptsize 0.0797 & \scriptsize 0.0454 & \scriptsize 0.0219 & \scriptsize 0.2363 & \scriptsize 0.0526 & \scriptsize 0.2432 & \scriptsize 0.8394 \\ \hline
    \end{tabular}
    \caption{arXiv Summarization Dataset Quantitative Evaluation Metrics}
    \label{table:label_placeholder}
\end{table}
\subsection{Qualitative Evaluation  }
To further evaluate the superiority of the proposed solution, we invited 19 human subjects to conduct qualitative assessments. Specifically, we used a 7-point Likert scale to evaluate the quality of text summaries generated by the frameworks under the guidance of Doubao, GPT, Claude, Zhipu, Mistral-7b, and ChallengeMe on the arXiv Summarization Dataset, BillSum, and CNN/Daily Mail datasets. The higher the score, the higher the quality of the generated text. The experimental results are shown in the table below. 
\begin{table}[h]
\setlength{\tabcolsep}{1.75pt}
    \centering
    \begin{tabular}{ccccccccccc}
\hline
        \textbf{} & \textbf{\scriptsize Doubao} & \textbf{\scriptsize GPT} & \textbf{\scriptsize Claude} & \textbf{\scriptsize Zhipu} & \textbf{\scriptsize Mistral-7b} & \textbf{\scriptsize 8.0} & \textbf{\scriptsize 8.2} & \textbf{\scriptsize 8.4} & \textbf{\scriptsize 8.6} & \textbf{\scriptsize 8.8} \\ \hline
        \scriptsize  arXiv & \scriptsize 4.5 & \scriptsize 4.7 & \scriptsize 4.7 & \scriptsize 4.4 & \scriptsize 4.5 & \scriptsize 4.4 & \scriptsize 5.5 & \scriptsize 5.7 & \scriptsize 5.6 & \scriptsize 5.9 \\ 
        \scriptsize  BillSum & \scriptsize 4.5 & \scriptsize 4.5 & \scriptsize 4.3 & \scriptsize 3.3 & \scriptsize 4.8 & \scriptsize 5.1 & \scriptsize 5.8 & \scriptsize 5.6 & \scriptsize 5.9 & \scriptsize 5.9 \\ 
        \scriptsize CNN & \scriptsize 4.9 & \scriptsize 4.5 & \scriptsize 5.2 & \scriptsize 4.7 & \scriptsize 4.9 & \scriptsize 5.8 & \scriptsize 5.8 & \scriptsize 6.1 & \scriptsize 6.2 & \scriptsize 6.2 \\ \hline
    \end{tabular}
    \caption{Qualitative Evaluation Metrics}
    \label{table:label_placeholder}
\end{table}

As shown in the table above, the version with a threshold of 8.8 achieved the best performance on the arXiv Summarization Dataset, BillSum, and CNN/Daily Mail datasets, with scores of 6.9, 5.9, and 6.2 respectively, significantly higher than the five baselines participating in the experiment. This indicates that the method proposed in this paper has significant superiority and can generate text summaries of higher quality compared to existing general large models. 
\section{Discussion and Conclusion }

\subsection{Human-AI Parallels: Fueling Next-Gen Research  }
With the rapid development of AI, the relationship between humans and AI has become increasingly close. However, existing research indicates that there are differences between AI and humans in understanding the real world and cognitive behavior \citep{konar2018artificial}. Human cognition is a complex and dynamic process involving multiple aspects such as perception, memory, emotion, and logical reasoning, and its cognitive abilities are formed through long-term biological evolution and individual experience accumulation. When facing problems, humans can flexibly use intuition, emotion, and past experiences to make judgments and decisions, and this cognitive style has high adaptability and flexibility. For example, when facing complex social situations, humans can make appropriate responses based on the perception and understanding of others' emotions. The cognitive behavior of AI large models mainly depends on the data they are trained on and the algorithms. By learning from a vast amount of data, they establish complex pattern recognition and prediction capabilities. AI demonstrates extremely high efficiency and accuracy in processing data and performing tasks, but when facing situations beyond its training scope, it often encounters difficulties in understanding and adaptation.

By comparing the cognitive patterns of humans and AI, and examining their similarities and differences, such as the flexibility and adaptability of human cognition, albeit with relatively lower efficiency; while AI excels in efficiency and accuracy, but lacks the flexibility and emotional understanding of humans. These differences provide important insights for the research of new-generation technologies. Future technological development should strive to combine the cognitive strengths of humans with the powerful processing capabilities of AI, to develop smarter, more flexible, and efficient artificial intelligence systems. For instance, introducing technologies such as affective computing and brain-like intelligence can enable AI to better understand and simulate human cognitive processes, thereby playing a role in a wider range of fields.
\subsection{Exploring AI Optimization Strategies through Cognitive Mechanisms}
In the exploration of AI optimization solutions, adversarial learning has become one of the key research directions. From the perspective of cognitive mechanisms, the human thinking process has a high degree of flexibility and creativity, and can make inferences and innovations based on limited data and experience. In contrast, the cognitive behavior of AI large models mainly relies on data-driven statistical learning, lacking the reasoning ability of humans based on theories and hypotheses. This difference provides important insights for AI optimization: by simulating an adversarial environment, AI models can enhance their robustness and adaptability in the process of continuous challenge and being challenged. By exploiting the model's own weaknesses to generate adversarial samples, it helps the model learn more comprehensive and complex feature representations. Adversarial learning provides new ideas for the performance improvement of AI models. By simulating human adaptability and creativity, AI can demonstrate stronger robustness and flexibility in complex environments. Future research should further combine the advantages of human cognition to develop more innovative and interpretable AI systems.
\subsection{Limitations \& Futurework}
Despite the valuable conclusions and findings obtained from the experiments, this study still has certain limitations. First, the case studies in the experiments focused on the text summarization task. Although comparative studies were conducted on multiple datasets and AI models, research on more tasks will further enhance the reliability of the research conclusions. In addition, during the experimental process, we selected the 4th generation models (based on the GPT series as the standard) for comparative studies. This is because they represent the most advanced large model solutions currently available. However, with the continuous iteration of technology, research on larger-scale parameters and more modal tasks will further increase value. Moreover, in future research, conducting studies with a larger number of participants and exploring variables that potentially affect cognitive levels, such as gender and age, will help further explore the similarities and differences between AI and humans and obtain inspiring insights, thereby constructing more advanced models.

\subsection{Conclusion}

Inspired by contrastive learning, this paper constructs an adversarial prompt framework named ChallengeMe, to tackle the limitations of large language models in content generation, such as hallucination and non-specific content. By conducting analysis using text summarization tasks as case studies, we compared the proposed framework's performance with current advanced large language models in three public text summarization datasets, and verifying the advancement and superiority of the proposed framework through quantitative, qualitative, and ablation experiments. The results and findings reveals potential ideas for the future evolution of large models.

%%%%%%%%%%%%%%%%%%%%%%%%%%%%%%%%%%%%%%%%%%%%%%%%%%%%%%%%%%%%%%%%%%%%%%%%%%%%%%%%%%%%%%%%%%%%%%%%%%%%%%%%%%%%%%%%%%%%%

% \nocite{ChalnickBillman1988a}
% \nocite{Feigenbaum1963a}
% \nocite{Hill1983a}
% \nocite{OhlssonLangley1985a}
% % \nocite{Lewis1978a}
% \nocite{Matlock2001}
% \nocite{NewellSimon1972a}
% \nocite{ShragerLangley1990a}

\printbibliography

\end{document}